\documentclass{article}

\usepackage{spconf,amsmath,graphicx}

\usepackage{amssymb}
\usepackage{dsfont}
\usepackage{amsfonts}
\usepackage{mathrsfs}

\usepackage{setspace}
\usepackage{xspace}

\usepackage{ stmaryrd }
\usepackage{lineno}
\usepackage{color}
\usepackage{url}
\usepackage{ stmaryrd }

\usepackage{amssymb,color} 
\usepackage{algorithm}
\usepackage{algpseudocode}

\usepackage{soul}

\newcommand{\Rr}{\mathds{R}}
\newcommand{\Cr}{\mathds{C}}

\newcommand{\R}{P}

\newcommand{\YYY}{\mathbf{Y}} 
\newcommand{\AAA}{\boldsymbol{\Psi}_{\mathbf{X}}} 
\newcommand{\BBB}{\boldsymbol{\Psi}_{\mathbf{Y}}} 
\newcommand{\ZZZ}{\mathbf{Z}}

\newcommand{\TT}{T}
\newcommand{\TTPrim}{T'}

\newcommand{\eg}{\textit{e.g.}, }
\newcommand{\ie}{\textit{i.e.}, }

\DeclareMathOperator*{\argmin}{arg\,min}

\newtheorem{proposition}{\it \bf Proposition}

\title{Generalized Kernel-Based Dynamic Mode Decomposition}

\name{Patrick H\'eas$^{1,2}$, C\'edric Herzet$^{1,2}$ and Benoit Comb\`es$^{1,3}$}
\address{$^1$INRIA, $^2$IRMAR, $^3$IRISA,  Univ. Rennes, Campus de Beaulieu, France}
\begin{document}
%
\maketitle
\begin{abstract}
Reduced modeling in high-dimensional reproducing kernel Hilbert spaces offers the opportunity to approximate efficiently non-linear dynamics. In this work, we devise an algorithm based on low rank constraint optimization and kernel-based computation that generalizes a recent approach called ``kernel-based dynamic mode decomposition". This new algorithm is characterized by a gain in approximation accuracy, as evidenced by  numerical simulations, and in computational complexity.
\vspace{-0.cm}

\end{abstract}
\begin{keywords}
 Reduced modeling, kernel-based methods,   low-rank approximations,  non-linear dynamics \vspace{-0.05cm}
\end{keywords}

\section{Introduction} \vspace{-0.125cm}

In this paper, we consider the problem of efficiently approximating trajectories $x_t(\theta) \in  \Rr^p$, for different initial conditions $\theta$ from the following high dimension system:
\begin{align}\label{eq:model_init} 
 \left\{\begin{aligned}
& x_{t}(\theta)= f_t(x_{t-1}(\theta)) , \quad t=2,\ldots,\TT,\\
&x_1(\theta)={\theta},
\end{aligned}\right. \vspace{-0.6cm}
\end{align} 
\noindent
where $f_t:\Rr^p \to \Rr^p$ is an arbitrary function whose direct evaluation is time consuming when $p$ is large. 

Dynamic Mode Decomposition~\cite{Chen12,Jovanovic12,Tu2014391} is a popular framework for this purpose and relies on efficient linear approximations of the trajectories of \eqref{eq:model_init}. It has been extended to the approximation of non-linear behaviors using a decomposition known as extended DMD (EDMD) \cite{williams2015data,williams2014kernel,Lusch2018DeepLF}.  Basically,  DMD and EDMD  are identical,  except that the latter first immerses the trajectory through a non-linear mapping $\Psi$ in a space exhibiting better approximation capabilities.   
More explicitly,
let $\Psi: \Rr^p \to \mathcal{H}$, where  $\mathcal{H}$  is a Hilbert space  endowed with the inner product  $\langle \cdot ,\cdot \rangle_{\mathcal{H}}$  and the induced norm $\| \cdot \|_\mathcal{H}$. 
EDMD approximates system \eqref{eq:model_init} by:
 \begin{align}\label{eq:model_koopman_approxObservable0} 
 \left\{\begin{aligned}
&\eta_t(\theta)= \hat A_k \eta_{t-1}(\theta), \quad t=2,\ldots,{\TT},\\
&\eta_1(\theta)=\Psi({\theta}),
\end{aligned}\right. \vspace{-0.4cm}
\end{align} 
where $\hat A_k : \mathcal{H} \to \mathcal{H}$ is a  linear  operator   of rank $ \le k$, satisfying some optimality criterion (specified later),  yielding an approximation of the state $x_{\TT}(\theta)$ by an inverse mapping \vspace{-0.2cm}
\begin{align}\label{eq:model_koopman_approxObservable} 
\tilde x_{{\TT}}(\theta)=\Psi^{-1}(\eta_{\TT}(\theta)).\vspace{-0.3cm}
\end{align}

 In this paper, we will focus on reduced models of the form  \eqref{eq:model_koopman_approxObservable0}-\eqref{eq:model_koopman_approxObservable} and  where   $\dim(\mathcal{H})\gg p$ (including $\dim(\mathcal{H})=\infty$). Such an embedding is appealing due to the ability of high-dimensional Hilbert spaces to linearize   differential equations~\cite{koopman1931hamiltonian,kowalski1991nonlinear,mezic2004comparison}.  
 To obtain a ``good'' trade-off between accuracy and complexity of the reduced model,  one needs to accomplish two challenging  tasks: {\it i)} learn a tractable representation of a low-rank operator $\hat A_k$ yielding an accurate approximation of the form \eqref{eq:model_koopman_approxObservable0}-\eqref{eq:model_koopman_approxObservable}, {\it ii)}   build a  low-complexity algorithm to compute  $\tilde x_{\TT}(\theta)$ satisfying  \eqref{eq:model_koopman_approxObservable0}-\eqref{eq:model_koopman_approxObservable} for a given $\theta$.

State-of-the-art methods \eg~\cite{Tu2014391,heas2017optimal,hemati2017biasing} involve a complexity in $\dim(\mathcal{H})$ and thus are  non-efficient in  high-dimensional settings.  In parallel, authors in \cite{williams2014kernel} have introduced an efficient  algorithm to compute \eqref{eq:model_koopman_approxObservable}  for any map $\Psi$ related to a reproducing kernel Hilbert space (RKHS) \cite{steinwart2006explicit}.
  This algorithm known as kernel-based DMD (K-DMD) enjoys  an advantageous complexity  linear in $p$ and independent of $\dim(\mathcal{H})$ but relies on restrictive assumptions. 
  
   In this work, we propose a new algorithm dubbed ``generalized kernel-based DMD (GK-DMD)'' that generalizes   K-DMD to less restrictive assumptions, while being characterized by a gain in computational complexity and approximation accuracy, as evidenced by our numerical simulations. \vspace{-0.125cm}
   
 

\section{Problem and Existing Solutions}\label{sec:StateOfTheArt}\vspace{-0.125cm}

\subsection{The Reduced Modeling Problem}
Let  $\mathcal{B}(\mathcal{V},\mathcal{U})$ denote the class of linear bounded operators from $\mathcal{V}$ to  $\mathcal{U}$ and let
$\mathcal{B}_k(\mathcal{V},\mathcal{U}) = \{  M \in \mathcal{B}(\mathcal{V},\mathcal{U}): \textrm{rank}(M)\le k\}$. 
 In this work, we consider a data-driven approach: the reduced model is learnt from a set of representative trajectories $\{x_t(\vartheta_i)\}_{t=1,i=1}^{\TTPrim,N}$ of the high-dimensional system corresponding to  $N$ initial conditions $\{ \vartheta_i\}_{i=1}^N$ (with $\TTPrim$ possibly different from $\TT$).  We are interested in the design of an algorithm computing for any $\theta \in \Rr^p$ the approximation $\tilde x_{\TT}(\theta)$ using a reduced model of the form   \eqref{eq:model_koopman_approxObservable0}-\eqref{eq:model_koopman_approxObservable} and defined as follows.

%
%
%

$\bullet$ {\bf Low-rank operator}.  The low-rank linear operator $\hat A_k$ is identified  to  a solution of the constrained optimization problem\vspace{-0.3cm}
\begin{align}\label{eq:prob1} 
		A_k^\star \in &\argmin_{A\in \mathcal{B}_k(\mathcal{H},\mathcal{H})} \|\BBB -A  \AAA \|_{\mathcal{HS}},\vspace{-0.2cm}
		\end{align} 
where  $\|\cdot\|_{\mathcal{HS}}$ refers to the Hilbert-Schmidt norm and where  operators $\AAA,\, \BBB \in \mathcal{B}(\Rr^m,\mathcal{H}) $, with $m=N(\TTPrim-1)$,  are defined for any $w \in  \Rr^m$ as the linear combinations 
$
 \AAA w = \sum_{i,j=1}^{N,\TTPrim-1} \Psi(x_{j}(\vartheta_i)) w_{(\TTPrim-1)(i-1)+j} $ and  
$ \BBB w=  \sum_{i,j=1}^{N,\TTPrim-1} \Psi(x_{j+1}(\vartheta_i)) w_{(\TTPrim-1)(i-1)+j}.$ These combinations 
involve the training data set $\{x_{j}(\vartheta_i)\}_{i,j=1}^{N,\TTPrim}$ , where the $i$-th component of a vector is denoted by  subscript $i$.
 Operator \eqref{eq:prob1} is a generalization of the solution of the minimization problem in \cite{williams2015data,williams2014kernel}, subject to a low-rank constraint as in~\cite{Chen12,Jovanovic12}.

$\bullet$  {\bf Minimum distance estimation}. The inverse map \eqref{eq:model_koopman_approxObservable} is defined as  a minimum distance estimate \vspace{-0.1cm}
\begin{align}\label{eq:defInverse2}
 \Psi^{-1}(\eta) &\in
\arg\min_{z \in  \Rr^p}\|{ \eta}-{\Psi(z)}\|_{\mathcal{H}}.\vspace{-0.3cm}
\end{align}

$\bullet$   {\bf Low-complexity}.  The algorithm's complexity is  independent of  $\dim(\mathcal{H})$ and the simulated trajectory length~$\TT$.


Moreover, in order to enable the independence in ${\TT}$, we will assume all along this work  that $\mathcal{H}$ is separable and that $A_k^\star $ is  diagonalizable.  These  assumptions enable  to evaluate  recursion~\eqref{eq:model_koopman_approxObservable0}  independently of the trajectory length ${\TT}$. 
Explicitly, let  $\{\xi_i\}_{i\in \mathbb{N}}$ and   $\{\zeta_i\}_{i\in \mathbb{N}}$ be   bases of $\mathcal{H}$ associated to the left and right eigen-vectors of  $A_k^\star $, \ie $\xi_i A_k^\star   = \lambda_i \xi_i$ and $A_k^\star  \zeta_i = \lambda_i \zeta_i$  for $i\in \mathbb{N}$, where $\{\lambda_i\}_{i \in  \mathbb{N}}$ is the related sequence of eigen-values sorted by decreasing magnitude. The finite rank of operator $A_k^\star $ and the bi-orthogonality of the left and right eigen-vectors yield  
$A_k^\star  \Psi = \sum_{i=1}^{k} \lambda_i\langle\xi_i, \Psi \rangle_{\mathcal{H}}   \zeta_i.$
Using the notation   $  \varphi_i(\theta) = \langle \xi_i, \Psi(\theta) \rangle_{\mathcal{H}}$, \eqref{eq:model_koopman_approxObservable} then  becomes \vspace{-0.2cm}
\begin{align}\label{eq:koopman1}
 \tilde x_{\TT}(\theta)&=\Psi^{-1} ( \sum_{i=1}^{k}\nu_{i,{\TT}}\zeta_i ),\quad 
\nu_{i,{\TT}} =  \lambda_i^{{\TT}-1}  \varphi_i(\theta).\vspace{-0.4cm}
\end{align}  

\subsection{Two Existing Solutions}
In the following, we discuss two existing methods which will serve as   ingredients for our GK-DMD algorithm. 

\textbf{Optimal but Intractable}. Reduced model \eqref{eq:koopman1} with $A_k^\star $  given by \eqref{eq:prob1} is referred to as \textit{low-rank EDMD}. 
  A generalization of \cite[Theorem 4.1]{HeasHerzet17}   to  separable infinite-dimensional Hilbert spaces  provides  a closed-form expression of operator $A_k^\star$~\cite{HeasHerzet18Maps}:
 a solution of  problem \eqref{eq:prob1} for arbitrary value of $k$ is\vspace{-0.15cm}
\begin{align}\label{eq:solAkOpt}
 A^\star_k=\mathbb{P}_{\ZZZ^k}\BBB \AAA^\dagger,\vspace{-0.25cm}
\end{align}
with the orthogonal projector $\mathbb{P}_{\ZZZ^k}=\hat \R_k \hat \R_k^*$.  We use short-hand SVD notations\footnote{We will use  the short-hand SVD notation for $M \in \mathcal{B}(\mathcal{V},\mathcal{U})$ : $M  =  U_M  \Sigma_M  V_M^*,$ where $U_M\in \mathcal{B}(\Cr^m,\mathcal{U})$,  $\Sigma_M\in \mathcal{B}(\Rr^m,\Rr^m)$ and $ V_M^*\in \mathcal{B}(\mathcal{V},\Cr^m)$ are defined 
for any vector $w\in  \mathcal{V},\, s \in \Cr^m$ as
$
 U_M s = \sum_{j=1}^{m} u_j^M s_{j}, \quad  (V_M w)_i = \langle v_i^M, w \rangle_{\mathcal{V}}\quad \textrm{and} \quad
( \Sigma_M s)_i=  \sigma_i^M s_i.
 $}
in order to define the operator  $\hat \R_k \in  \mathcal{B}(\Rr^k, \mathcal{H}):  w \to  \sum_{i=1}^k u^{\ZZZ}_i w_i $ with   $\ZZZ\in \mathcal{B}(\Rr^m, \mathcal{H})$ as \vspace{-0.15cm}
  \begin{align}
  \label{eq:Z} 
  \ZZZ= \BBB \mathbb{P}_ {\AAA^*}.\vspace{-0.2cm}
  \end{align}
   It can be shown that if  $k \ge m$, the solution of \eqref{eq:prob1} boils down to   the solution 
   of the unconstrained problem~\cite{Tu2014391}
$  \hat A^{\ell s}_k=\BBB  \AAA^\dagger.$
However, it remains to propose a tractable algorithm to build and evaluate reduced model  \eqref{eq:koopman1} from the closed-form,  but potentially infinite-dimensional,  solution $A_k^\star$.

\textbf{Tractable but Restrictive.}
To tackle the high-dimensional setting  $\dim(\mathcal{H})  \gg p$,  authors propose to consider    in their seminal work  a specific class of  mapping $\Psi$  from $\Rr^p$ to $\mathcal{H}$~\cite{williams2014kernel}. They 
assume $\mathcal{H}$ to be a RKHS~\cite{steinwart2006explicit}. Such a   space  of functions on $\Rr^p$ is uniquely determined by the choice of a  symmetric positive definite kernel   
$
h\,:\,\Rr^p \times \Rr^p \to \Rr,
$
such that  $\langle \Psi(z), \Psi(y) \rangle_{\mathcal{H}}= h(y,z)$ with $z, y \in \Rr^p$. 
 The advantage of such a construction is that the kernel trick~\cite{bishop2006pattern}
 can  be used to compute inner products in the RKHS $\mathcal{H}$ with a complexity equal to that required for the evaluation of the function $h$, which is in general independent of $\dim(\mathcal{H})$.
More specifically, their method called K-DMD uses the kernel trick  to evaluate inner products with  eigen-vectors of $  \hat A^{\ell s}_k$.
 Assuming that the complexity for the evaluation of the kernel is $\mathcal{O}(p)$,   the overall complexity of the K-DMD algorithm is independent of $\dim(\mathcal{H})$ and ${\TT}$, which may be  efficient for  $\dim(\mathcal{H})  \gg p$. 

 However, as proposed in  \cite{williams2014kernel},  K-DMD computes an approximation of reduced model \eqref{eq:koopman1} under restrictive assumptions.  In particular the four following assumptions are needed:  
{\it  i)}  $A^\star_k=\hat A^{\ell s}_k$, \ie  the low-rank constraint in \eqref{eq:prob1} is ignored; 
{\it  ii)} the operator $\AAA$ is full-rank;
{\it  iii)} $\Psi^{-1}$ is linear;
{\it  iv)}   the  $ \Psi^{-1} \zeta_j$'s  belong to the span of $\YYY$, 
where the  elements in the set $\{x_{t+1}(\vartheta_i)\}_{t=1,i=1}^{\TTPrim-1,N}$ define the columns $\{y_i\}_{i=1}^{m}$ of matrix $\YYY\in \Rr^{p \times m}$.
\vspace{-0.2cm}

 \section{A Generalized Kernel-Based Algorithm} \label{sec:okDMD} \vspace{-0.cm}

 \vspace{-0.2cm} \subsection{The GK-DMD Algorithm}  \vspace{-0.cm} 
Our generalized kernel-based  algorithm,  called GK-DMD, is exposed in Algorithm~\ref{algo:3}. It computes the low-rank reduced model \eqref{eq:koopman1} for $\mathcal{H}$ being  an RKHS, with a complexity independent of $\dim(\mathcal{H})$ and $\TT$ and is relieved from the  assumptions made in K-DMD.  
 As for K-DMD, the GK-DMD exploits the kernel-trick in step 1) and 6), and resorts to an analogous computation of eigen-functions in step 7). 
The main innovation  in comparison to the latter state-of-the-art algorithm is that GK-DMD computes reduced model \eqref{eq:koopman1} based on the exact solution \eqref{eq:solAkOpt} of problem~\eqref{eq:prob1}.  To enable the reduced model to be tractable  with  the  solution~\eqref{eq:solAkOpt}, GK-DMD relies on the two following original results:

$\bullet$  the right and left eigen-vectors of the optimal operator  $ A^\star_k$ belong to a low-dimensional sub-space of $\mathcal{H}$; their low-dimensional representations are tractable  and computed in steps 1) to 5) relying on the kernel function; 

 $\bullet$ the inverse map  defined in \eqref{eq:defInverse2} involves a distance minimization problem in $\mathcal{H}$; 
 Taking advantage that, in  reduced model~\eqref{eq:koopman1}, the argument of the inverse belongs to a low-dimensional subspace of $\mathcal{H}$, the high-dimensional minimization problem boils down to a tractable $p$-dimensional optimization problem computed in step 8). 
 
 These two results are detailed in  Section~\ref{sec:preuveOptimal}. 
 
  \vspace{-0.cm} 
 \begin{algorithm}[h]
 \begin{algorithmic}[0]
\State $\bullet$ \textbf{Off-line.} {\bf Inputs:}   $x_t(\vartheta_i)$'s 

\begin{algorithmic}[0]
\State 1) Compute matrices   $\AAA^*  \AAA$, $\BBB^*  \BBB$, $\BBB^*  \AAA  $  in $\Rr^{m \times m}$  with the kernel trick.
\State 2) Compute $(V_{\AAA} , \Sigma_{\AAA})$ by eigen-decomposition of  $\AAA^*   \AAA$. 
%
\State 3) Compute $(V_{\ZZZ} , \Sigma_{\ZZZ})$ by eigen-decomposition of  $\ZZZ^* \ZZZ$ with  $\ZZZ$ given by \eqref{eq:Z}. 
\State 4) Compute the two matrices given  in  Proposition~\ref{prop:3.3}  and compute  their eigen-vector/eigen-value couples $\{ (\tilde \xi_i ,\tilde \lambda_{i}) \}_{i=1}^k$ and $\{ (\tilde \zeta_i ,\tilde \lambda_{i}) \}_{i=1}^k$.  
\State 5)  Rescale  $\tilde \zeta_i$'s so that $\tilde \zeta_i\, E\, \tilde \xi_i$=$1$ with
$
E$=$S_k\BBB^* \AAA R^*.$
\State \textbf{Outputs}: $R$, $S_k$, $ \tilde \xi_i$'s, $ \tilde \zeta_i$'s and $\tilde \lambda_i$'s
\end{algorithmic}
\end{algorithmic}
\begin{algorithmic}[0]
\State $\bullet$ \textbf{On-line.} \textbf{Inputs}:  off-line outputs and $\theta$
\begin{algorithmic}[0]
\State 6) Compute  $ \AAA^*  \Psi(\theta)$ in $\Rr^{m}$ with the kernel trick.
\State 7) Compute  eigen-functions  $\{ \varphi_i(\theta)\}_{i=1}^k$ defined as \vspace{-0.2cm}
\begin{align}\label{eq:eigfuncApprox0}
 \varphi_i(\theta)=  \langle \xi_i, \Psi(\theta) \rangle_{\mathcal{H}}= 
 \tilde \xi_i^* R \AAA^*  \Psi(\theta).\vspace{-0.4cm}
\end{align}
\State 8) Compute  $ \tilde x_{\TT}(\theta) $ solving \eqref{eq:defInverse3}; 
\State \textbf{Output}: $\tilde x_{\TT}(\theta)$.
\end{algorithmic}
\end{algorithmic}
\caption{: GK-DMD   \label{algo:3}} 
\end{algorithm}
~\vspace{-0.9cm}\\  Let us meanwhile analyze the advantage of the GK-DMD algorithm in terms of computational complexity.
 Assuming that the complexity for the evaluation of the kernel is $\mathcal{O}(p)$, the overall complexity of the proposed algorithm scales in  $\mathcal{O}(m^2(m+p))$, just as for  K-DMD.  We remark that this  complexity is independent of ${\TT}$ thanks to the eigen diagonalization of $A^\star_k$,  and independent of $\dim(\mathcal{H})$ due to the use of the kernel-trick  in the first and last steps of the algorithm. 
 Nonetheless,  reduced modeling is very concerned by the on-line computational cost, \ie complexity of computation steps depending on the input $\theta$.  As  $k\le m \le p$ and typically $k\ll p$, GK-DMD is attractive  by its   on-line complexity  in $\mathcal{O}(m^2k+mp )$, \ie it scales linearly with respect to the dimension of the reduced model $k$ or the ambient dimension $p$,  in comparison to   $\mathcal{O}(m^2p)$ operations for K-DMD.
 Indeed, the matrix-vector product  $ \AAA^*  \Psi(\theta)$  in step 6)  and  the inversion in step 8) are both computed in $\mathcal{O}(pm)$ operations, while eigen-functions in step 7) require  $\mathcal{O}(m^2k)$  operations. 
  \vspace{-0.3cm}



\subsection{Ingredients for Optimality}  \label{sec:preuveOptimal} \vspace{-0.2cm}

 In the two next sections, we prove that GK-DMD computes reduced model \eqref{eq:koopman1} based on the optimal solution \eqref{eq:solAkOpt} of~\eqref{eq:prob1}. \vspace{-0.3cm}
\subsubsection{Low-Dimensional Representation of $ A^\star_k$}\vspace{-0.2cm}

Steps 1) to 5) of our algorithm rely on the following proposition.  Let $\{ \xi_i \}_{i=1}^k$  and $\{ \zeta_i \}_{i=1}^k$   denote the left and right eigen-vectors  of $ A^\star_k$ associated to its at most $k$ non-zero eigen-values $\{ \lambda_i \}_{i=1}^k$. \vspace{-0.2cm}  

\begin{proposition}\label{prop:3.3} 
For $i=1,\ldots,k$, the left and right eigen-vectors  of $ A^\star_k$ and its eigen-values satisfy 
 \,\,$
\xi_i=U_{\AAA} \tilde \xi_i \textrm{,}\quad \zeta_i=\hat P_k \tilde \zeta_i \quad \textrm{and}\quad \lambda_i=\tilde \lambda_{i}
$ where
 $\{ (\tilde \xi_i ,\tilde\lambda_{i}) \}_{i=1}^k$ and  $\{ (\tilde \zeta_i ,\tilde \lambda_{i}) \}_{i=1}^k$ denote  respectively  the first  $k$ right eigen-vectors and   eigen-values of the  matrices 
$
R\,  \BBB^* \BBB\, S_k^*S_k \, \BBB^* \AAA\, R^* \in \Rr^{m \times m} $ and $  S_k\,  \BBB^* \BBB\, R^*\,R \, \AAA^* \BBB\, S_k^* \in \Rr^{m \times m},
$
with   $R= \Sigma_{\AAA}^\dagger V_{\AAA}^*$ and $ S_k= \textrm{diag}((\sigma^\ZZZ_{1})^{\dagger}\cdots (\sigma^\ZZZ_{k})^{\dagger}0\cdots 0) V_\ZZZ^* $. 
\vspace{-0.cm}
\end{proposition}
 
 \vspace{-0.cm}Proposition \ref{prop:3.3} gives    a decomposition of the left eigen-vectors of  $ A^\star_k$ given in~\eqref{eq:solAkOpt}. Its proof is detailed in \cite{Heas2020a}. We deduce from Proposition~\ref{prop:3.3} the closed-form $i$-th   eigen-function approximation $ \varphi_i(\theta)$ for  $i=1,\ldots, k$  at any point  $\theta \in \Rr^p$ given in  \eqref{eq:eigfuncApprox0}.
Moreover, this proposition  provides a closed-form decomposition for the $\zeta_i$'s, the right eigen-vectors of  $ A^\star_k$ and supplies the related eigen-values.
Thanks to  Proposition~\ref{prop:3.3}, the elements in $\{(\xi_i,\zeta_i,\lambda_i)\}_{i=1}^k$ issued from the eigen-decomposition of $A^\star_k$ (which correspond to  the parameters of  reduced model \eqref{eq:koopman1}) can be written in terms of their low-dimensional counterpart  $\{(\tilde \xi_i,\tilde \zeta_i,\tilde \lambda_{i})\}_{i=1}^k$ efficiently computed in the 5 off-line steps. 
 Note that  some simple algebraic calculus show that the normalization of the eigen-vectors   is ensured if $\tilde \zeta_i $ is rescaled as
$
\tilde \zeta_i ^*  E  \tilde \xi_i=1,
$
with $
E=S_k\BBB^* \AAA R^*.$\vspace{-0.3cm}

\subsubsection{Kernel-Based Inversion}\vspace{-0.2cm}

 The low-dimensional representation of eigen-vectors of $A^\star_k$  provided in Proposition~\ref{prop:3.3}  constitutes the main ingredient of the GK-DMD algorithm. However, to achieve the design of this algorithm, it  remains to provide a feasible manner to compute $\Psi^{-1}$ in \eqref{eq:koopman1}. Once more, the  idea  consists in relying  on the kernel trick in order to compute the inverse with a complexity independent of $\dim(\mathcal{H})$. 
 
 Using Proposition~\ref{prop:3.3}, we begin by rewriting  
 \eqref{eq:koopman1} in terms of  $\tilde \zeta_i$'s,  $\varphi_i(\theta)$'s and $\tilde \lambda_{i}$'s as\vspace{-0.1cm}
\begin{equation*}
\tilde x_{\TT}(\theta)
=\Psi^{-1}(\sum_{j=1}^k \hat \R_k \tilde \zeta_j  \tilde \lambda_{j}^{{\TT}-1} \varphi_j(\theta)) 
=\Psi^{-1}(\BBB g^{\theta,{\TT}}), \vspace{-0.4cm}
\end{equation*}
with 
$
g^{\theta,{\TT}}$=$S_k^* (\tilde \zeta_1 \cdots \tilde \zeta_k)  \begin{pmatrix}\tilde \lambda^{{\TT}-1}_{\ell,1} \varphi_1(\theta) &\cdots& \tilde \lambda^{{\TT}-1}_{\ell,k}\varphi_k(\theta)\end{pmatrix}^*
$  in $ \Rr^{m}$.
This equation implies the inverse of a linear combination of the $\Psi(y_i)$'s, where  $y_{i}=x_{t+1}(\vartheta_j)$ with $i=(\TTPrim-1)j+t$ for $j=1,\ldots, N$ and $t=1,\ldots, \TTPrim-1$. From \eqref{eq:defInverse2}, we  rewrite the inverse of the linear combination in terms of scalar products in $\mathcal{H}$ computable using the kernel trick, 
 \ie given the kernel $h$, \vspace{-0.3cm}
 \begin{align}\label{eq:defInverse3}
\tilde x_{\TT}(\theta)
\in \arg\min_{z\in \Rr^p} \left(  h(z,z)-2\sum_{i=1}^m  g_i^{\theta,{\TT}} {h(y_i,z)} \right). \vspace{-0.3cm}
\end{align}
The minimizer  can be computed (up to some accuracy) using standard  optimization methods with a complexity independent  of $\dim(\mathcal{H})$. Moreover, the gradient of the objective is in general  closed-form, which enables the use of efficient  large-scale optimization techniques such as limited memory quasi-newton methods~\cite{nocedal2000numerical}. In this case, the complexity to compute the inverse is  linear in $p$.\vspace{-0.2cm}

  \begin{figure}[!t]
\centering
\hspace{-0cm}\begin{tabular}{cccc}
\hspace{-1.cm} \includegraphics[height=0.135\columnwidth]{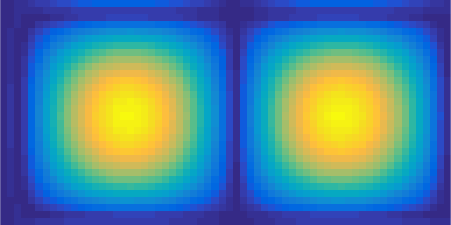}  
 &\hspace{-0.45cm}   \includegraphics[height=0.135\columnwidth]{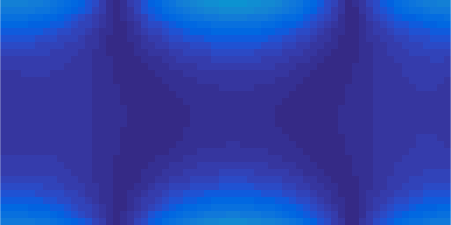} &
\hspace{-0.45cm}   \includegraphics[height=0.135\columnwidth]{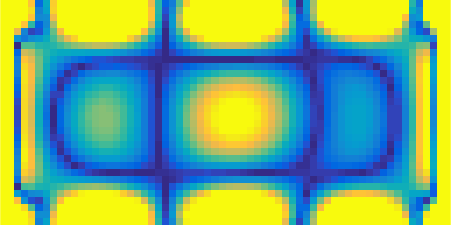}
&\hspace{-0.45cm}  \includegraphics[height=0.135\columnwidth]{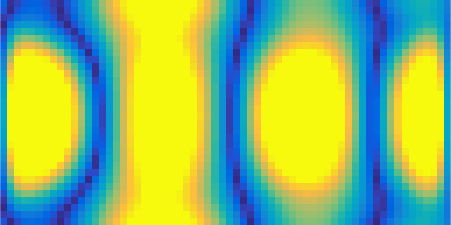} \\
 \multicolumn{2}{c}{\hspace{-1.75cm} \includegraphics[height=0.475\columnwidth]{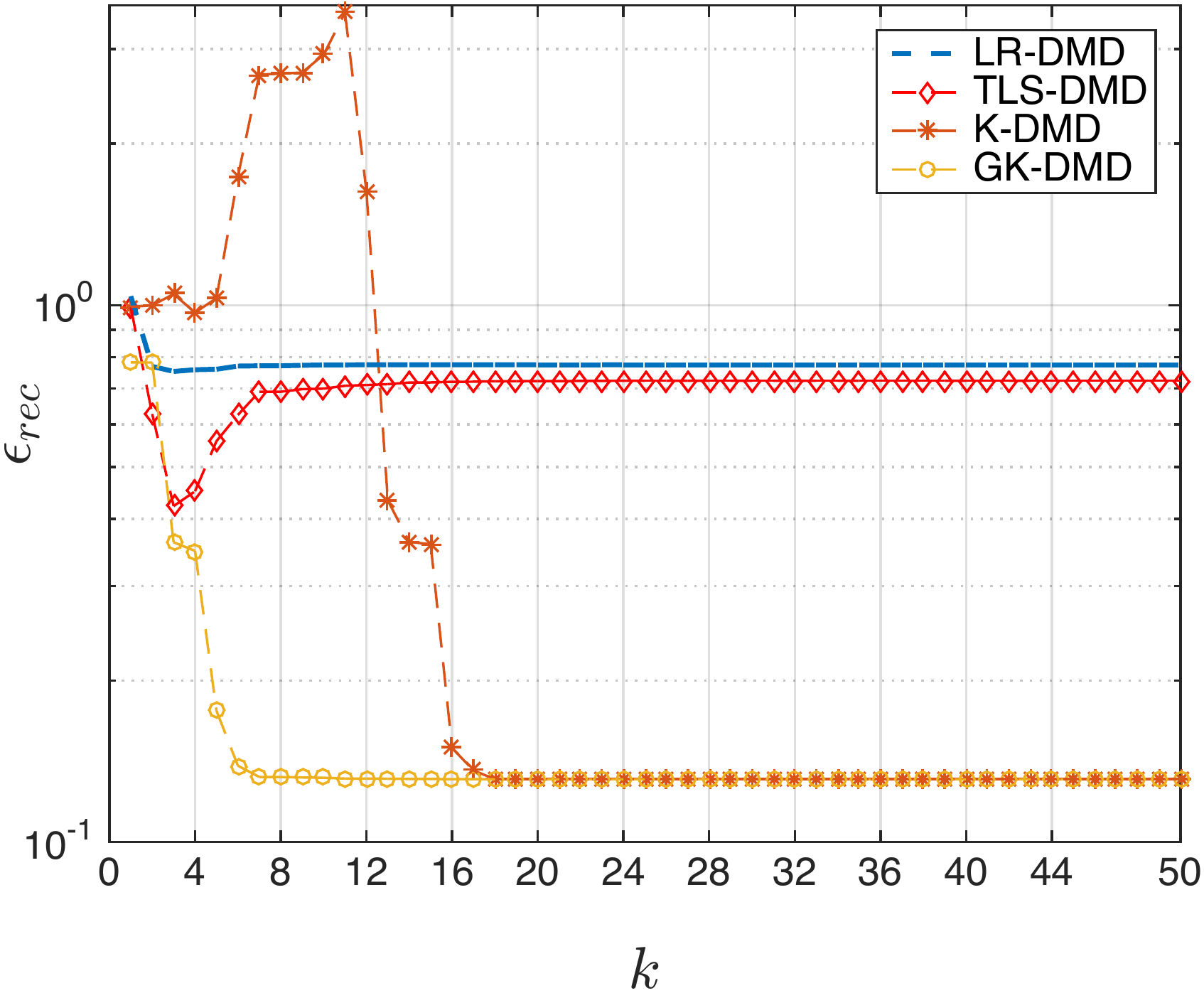}}&\multicolumn{2}{c}{\hspace{-0.5cm}  \includegraphics[height=0.475\columnwidth]{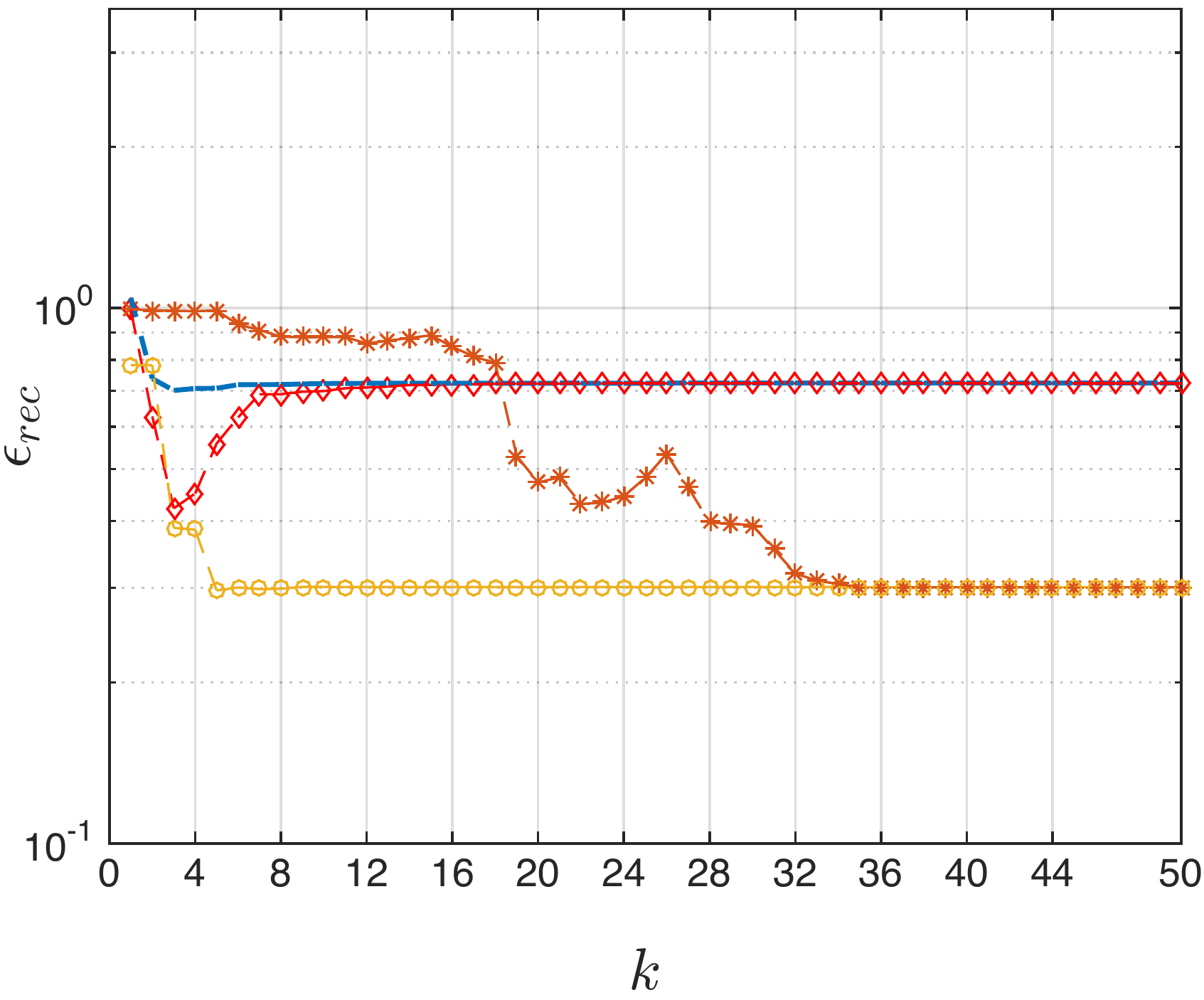}}\vspace{-0.5cm} \\

\vspace{-0.4cm}
\end{tabular}
\caption{\small Above:  two maps of  absolute vorticity (left, colors  in the range $[0,10^{-3}]$) and associated temperature fields (right, colors  in the range $[0,10^{-6}]$) for Rayleigh-B\'enard convection. Below:  reconstruction error $\epsilon_{rec}$  as a function of     rank~$k$  for GK-DMD and K-DMD with Gaussian (left) and polynomial  (right) kernels. \vspace{-0.cm}\label{fig:new0}\label{fig:1}}\vspace{-0.3cm}
\end{figure}

\begin{figure}[t!]
\centering
\hspace{-0.cm}
\hspace{-0.cm}\begin{tabular}{  c c c c}
\hspace{-1.cm}{\footnotesize LR-DMD}&\hspace{-0.45cm} {\footnotesize TLS-DMD}&\hspace{-0.45cm} {\footnotesize K-DMD}&\hspace{-0.45cm} {\footnotesize GK-DMD}\vspace{-0.cm}\\
\hspace{-1.cm}\includegraphics[height=0.135\columnwidth]{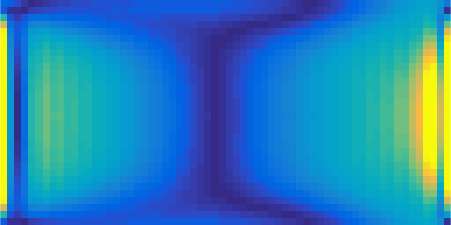}  &\hspace{-0.45cm} \includegraphics[height=0.135\columnwidth]{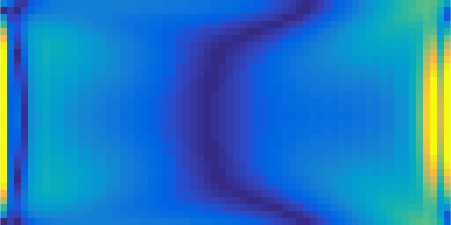}   &\hspace{-0.45cm} \includegraphics[height=0.135\columnwidth]{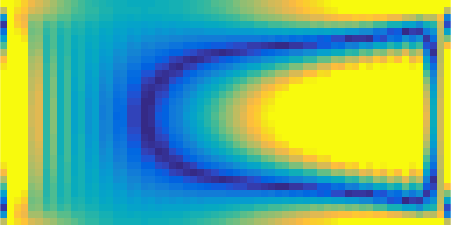} &\hspace{-0.45cm} \includegraphics[height=0.135\columnwidth]{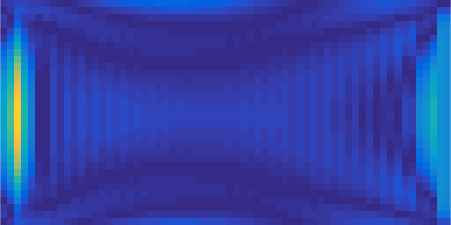}     \\
\hspace{-1.cm}\includegraphics[height=0.135\columnwidth]{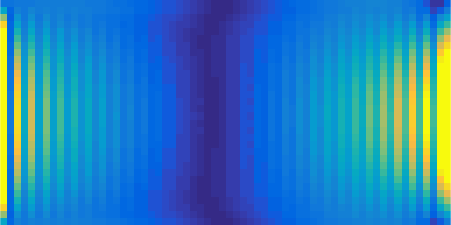}  &\hspace{-0.45cm} \includegraphics[height=0.135\columnwidth]{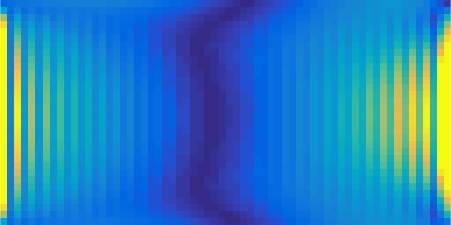}   &\hspace{-0.45cm} \includegraphics[height=0.135\columnwidth]{./Figures/XTLS15_38_b_error_crop.png} &\hspace{-0.45cm} \includegraphics[height=0.135\columnwidth]{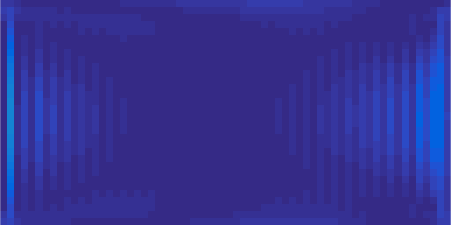}  \vspace{-0.15cm} \\
\end{tabular}
\caption{\small Reconstruction error maps produced for $k=5$ (above) and $k=15$ (below). Images  represent the absolute  vorticity (with colors in the range $[0,10^{-3}]$) of the   field ${ \tilde x_2( \theta)-x_{2}(\theta)}$ for a typical   $ \theta$. \vspace{-0.cm}\label{fig:new3}}\vspace{-0.3cm}
\end{figure}

\section{Numerical Simulations}\label{sec:numEval} \vspace{-0.2cm}

We assess four data-driven reduced modeling methods for  the approximation of   Rayleigh-B\'enard convection \cite{chandrasekhar2013hydrodynamic}, which is a standard benchmark model  in meteorology. Convection is driven by  two coupled  partial differential equations. 
After discretisation of these equations, we obtain  a discrete system with $p=4096$ for the evolution of  vorticity and temperature.

The benchmark algorithms are:
1) low-rank  DMD (LR-DMD) \cite[Algorithm 3]{HeasHerzet17},
2)  total-least-square  DMD (TLS-DMD) \cite{hemati2017biasing}, 
3)  kernel-based DMD (K-DMD) \cite{williams2014kernel}, 
4)  the proposed generalized kernel DMD (GK-DMD), \ie Algorithm \ref{algo:3}.
For the K-DMD and GK-DMD algorithms,   we use a quadratic   polynomial kernel  or  a Gaussian kernel with a standard deviation of $10$~\cite{bishop2006pattern}. 

     We study the evolution of     
  the {reconstruction error}   
 $
\epsilon_{rec}= \left(\sum_{j=1}^{N_{\theta}} \sum_{t=1}^{\TT-1}\frac{ \|\tilde x_2( x_{t}(\theta_j))-x_{t+1}(\theta_j)\|^2_2}{ \|x_{t+1}(\theta_j)\|^2_2}\right)^{1/2},
$
  with respect to the rank  $k$, for a set of initial conditions $\{\theta_j\}_{j=1}^{N_\theta}$. It measures the discrepancy
 between the true state $x_{t+1}(\theta_j)$ at time $t+1$ and the approximated state
 $\tilde x_2(x_{t}(\theta_j))$ predicted with the reduced model from the true state  at time $t$.

 The training data of size $m=90$ is set as follows: 10 initial
conditions $\vartheta_j$ are sampled from a uniform distribution on an
hyper-cube  in $\Rr^5$ parametrizing  solutions of the Lorenz attractor~\cite{Lorenz63};  then using $\vartheta_j$ to initialize  the dynamic model, we compute trajectories for $t=1,\ldots,10$ (resulting in 100 states $x_t(\vartheta_j)$). 
Examples of $x_t(\vartheta_j)'s$ are displayed in Figure~\ref{fig:1}.
 The test data  is set as the prolongation of the training data trajectories:  the 10 initial conditions  are  $\theta_j=x_{10} (\vartheta_j)$  and  trajectories
$x_t(\theta_j)$  for $t=1,\ldots,10$  are computed in the same way as for
the training data set.

 We first discuss the results  shown in Figure~\ref{fig:new0} for the Gaussian kernel.  Overall, we observe that GK-DMD outperforms almost everywhere the other methods.   While K-DMD and GK-DMD perform similarly  for  $k\ge 18$, for $k<18$ GK-DMD exhibits a clear gain in accuracy compared to the other methods  reaching almost a decade.   The  gain in accuracy  between  K-DMD and   GK-DMD  may be due to the fact that the GK-DMD  computes exactly  reduced model \eqref{eq:koopman1}, \ie considers  $A^\star_k$ instead of $\hat A^{\ell s}_k$.  
 Besides, as $\textrm{rank}(\AAA^*\AAA)=m$, \ie operator  $\AAA$ is full-rank,  a reasonable explanation for the    similar performances of the two kernel-based methods  in the case where  $k\ge 18$  is that the low-rank constraint becomes inactive
 (implying that $\hat A^{\ell s}_k=A^\star_k$),     $\Psi^{-1}$ is well approximated by a linear mapping and furthermore the $ \Psi^{-1} \zeta_j$'s are well represented in the span of $\YYY$. A lower value on the accuracy is reached around $k$ slightly greater than $5$, suggesting that only $5$ components  in $\mathcal{H}$ can be explained by a linear model. Similar results are obtained with a polynomial kernel. Nevertheless, the gain in accuracy is lower for polynomials, revealing that the reduced model performance is kernel-dependent.

Additionally, the performances of GK-DMD,  LR-DMD and TLS-DMD are comparable for $k < 4$. Nevertheless, the accuracy of  LR-DMD and TLS-DMD reaches a lower bound around $k \simeq 4$ and then 
deteriorates as $k$ increases  or reaches an asymptote, suggesting data overfitting. 

{To complement  this quantitative  evaluation, we  proceed to the visual inspection of the   spatial distribution of the error. Typical error maps  are shown in  Figure~\ref{fig:new3}. It displays the absolute vorticity  of the bi-variate error  field  ${ \tilde x_2( \theta)-x_{2}(\theta)}$ defined over the bi-dimensional grid, where $\tilde x_2(\theta)$ denotes the  approximation provided by  the algorithms for a given initial condition $\theta$. Error maps are displayed for two values of the dimension $k$.  The distribution of the error produced by K-DMD reveals that its chaotic behavior as $k$ increases is caused by errors in a wide range of scales. Error maps of the LR-DMD and TLS-DMD algorithms are very similar. Moreover they seem not to  involve significantly as $k$ increases, except for high frequency  revealed at $k = 15$.      The error maps for GK-DMD show that the decrease in error with respect to $k$ is related to  refinements occurring at increasingly finer scales.  }\vspace{-0.3cm}

  \section{Conclusion}\label{sec:conclusion}\vspace{-0.3cm}
We have presented a new algorithm for the tractable representation  of a linear low-rank operator characterizing dynamics embedded in a RKHS. By contrast to existing algorithms, it both exhibits a low computational complexity and requires mild assumptions.  Numerical simulations illustrate the gain in accuracy allowed by the proposed algorithm.\vspace{-0.3cm}

  \section*{Acknowledgements}\vspace{-0.2cm}
This work was supported by the French
Agence Nationale de la Recherche through the BECOSE Project.

\bibliographystyle{IEEEbib}
\bibliography{./bibtex}

\end{document}